\documentclass[letterpaper,journal]{IEEEtran}

\usepackage{amsmath,amsfonts,amssymb}
\usepackage{array}
\usepackage[caption=false,font=normalsize,labelfont=sf,textfont=sf]{subfig}
\usepackage{textcomp}
\usepackage{stfloats}
\usepackage{url}
\usepackage{verbatim}
\usepackage{graphicx}
\usepackage{cite}
\usepackage{xcolor}
\usepackage{multirow}
\usepackage{booktabs}
\usepackage{colortbl}
\usepackage{makecell}
\usepackage{bm}
\usepackage[hidelinks]{hyperref}

\definecolor{tabfirst}{rgb}{1, 0.7, 0.7}
\definecolor{tabsecond}{rgb}{1, 0.85, 0.7}
\definecolor{tabthird}{rgb}{1, 1, 0.7}

\begin{document}

\title{SceneCompleter: Dense 3D Scene Completion for Generative Novel View Synthesis}

\author{Weiliang Chen, Jiayi Bi, Yuanhui Huang, Wenzhao Zheng, and Yueqi Duan,~\IEEEmembership{Member,~IEEE}
\thanks{Weiliang Chen, Jiayi Bi, and Yueqi Duan are with the Department of Electronic Engineering, Tsinghua University, Beijing, China (e-mail: cwl24@mails.tsinghua.edu.cn; bijy22@mails.tsinghua.edu.cn; duanyueqi@tsinghua.edu.cn).}
\thanks{Yuanhui Huang and Wenzhao Zheng are with the Department of Automation, Tsinghua University, Beijing, China (e-mail: huangyh22@mails.tsinghua.edu.cn; wenzhao.zheng@outlook.com).}
\thanks{Corresponding author: Yueqi Duan.}
}

\maketitle

\begin{abstract}
Generative models have shown great promise for novel view synthesis (NVS) by leveraging strong image generation priors. However, existing approaches typically follow a 2D inpainting paradigm, first completing missing image regions and then performing 3D reconstruction. This strategy often causes geometry distortion and appearance drift, as 2D inpainting models cannot reliably infer the underlying 3D structure required for cross-view consistent generation.
In this paper, we propose \textbf{SceneCompleter}, a geometry-aware framework that reformulates generative NVS as dense 3D scene completion. Instead of hallucinating isolated 2D views, SceneCompleter jointly completes geometry and appearance through a geometry-appearance dual-stream diffusion model in a spatially aligned RGBD latent space. To provide holistic scene context, we further introduce a Scene Embedder that conditions generation on global semantic and stylistic information from reference images. The completed RGBD predictions are then aligned and integrated into an expandable 3D scene representation, enabling iterative and coherent scene completion. Extensive experiments on in-domain and out-of-distribution datasets demonstrate that SceneCompleter produces visually plausible and geometrically consistent novel views across diverse scenarios.
\end{abstract}


\section{Introduction}
\begin{figure*}[t]
    \centering
    \includegraphics[width=1\linewidth]{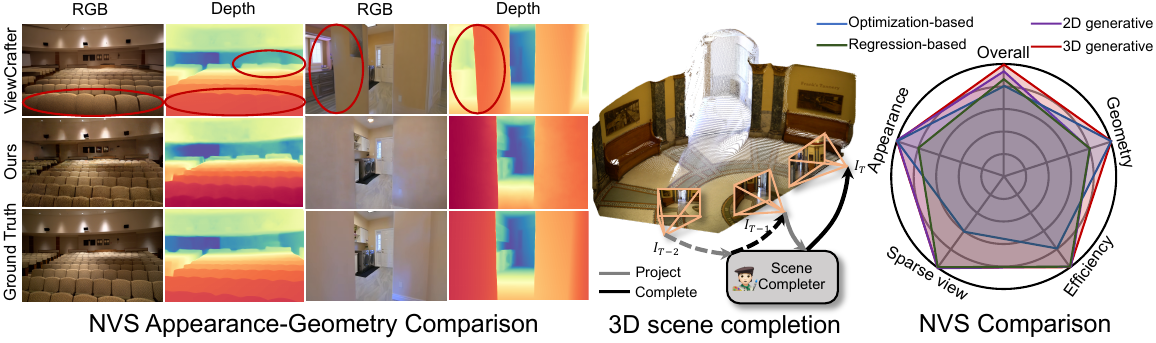}
    \caption{\textbf{SceneCompleter} explores 3D scene completion for generative novel view synthesis. By jointly modeling geometry and appearance, and incorporating geometry information into the generative process, SceneCompleter enables geometrically consistent and visually compelling novel view synthesis. With SceneCompleter, we can iteratively complete 3D scenes, ensuring both appearance and structure are accurately restored.}
    \label{fig:teaser}
\end{figure*}

Novel view synthesis (NVS) has gained significant attention in computer vision due to its broad applications in virtual reality~\cite{cho2019novelvirtual}, 3D content creation~\cite{long2024wonder3d,liu2023zero123,liu2024reconx,liu2024makeyour3d,liu2023sherpa3d,chen2024dreamcinema}, autonomous driving~\cite{lu2024drivingrecon,huang2024selfocc,zheng2024occworld,cao2023scenerf} and beyond. The core challenge lies in inferring 3D structure and appearance from limited views while generating plausible and visually coherent novel views. Despite achieving promising results with powerful differentiable 3D representations~\cite{mildenhall2021nerf, kerbl20233dgs}, optimization-based methods rely on dense multi-view inputs to dynamically search for the 3D structure, posing challenges in both efficiency and practical application. 

Recently, regression-based methods~\cite{charatan2024pixelsplat, chen2024mvsplat} have explored feedforward novel view synthesis, directly regressing pixel-aligned 3D representation parameters from sparse views. Despite showing promising results for nearby viewpoints through scene priors learned from large-scale training data, the ill-posed nature of the problem causes these methods to produce unsatisfactory artifacts and unrealistic geometry when applied to other viewpoints. Pioneered by Zero-1-to-3~\cite{liu2023zero123}, research~\cite{chung2023luciddreamer,yu2024viewcraftertamingvideodiffusion, liu2024reconx} has begun to explore generative novel view synthesis using powerful generative models. These methods typically involve synthesizing novel views with image or video generation models~\cite{ho2020ddpm,rombach2022ldm,peebles2023dit,xing2024dynamicrafter}, followed by estimating the 3D structure from the generated views for downstream reconstruction. However, the implicit nature of performing novel view synthesis in 2D pixel space makes it challenging for these methods to infer 3D structure, leading to distorted geometry. For example, in a projected incomplete view of a hall as shown in Figure~\ref{fig:teaser}, a 2D generative model might fill in overly smooth chair backs in the missing areas, while overlooking the armrests. This may happen because the armrests of the chairs occupy much less area than the backs, causing the model to interpolate the appearance of the chair backs in the missing areas. However, in 3D space, the small area of the chair armrests holds greater geometric significance, offering 3D clues for novel view synthesis.

In this paper, we introduce SceneCompleter, a novel framework that leverages dense 3D scene completion to enable geometry-consistent generative novel view synthesis. The core insight is that geometry information is crucial in generative novel view synthesis, as the model needs to infer the 3D structure of the scene and extrapolate to generate the missing area. Therefore, simultaneously modeling geometry and appearance is crucial for generative novel view synthesis. Specifically, we first extract the geometry and appearance clues from the reference view with a powerful stereo reconstruction model Dust3R~\cite{wang2024dust3r}. Then we design a Geometry-Appearance Dual-stream Diffusion model to perform generative novel view synthesis in 3D geometry-appearance space $(A, G)$. By jointly modeling geometry and appearance, this approach enables the generation of geometrically plausible novel views. Additionally, we introduce a Scene Embedder that encodes the overall scene information from the reference view to guide generation, which plays a crucial role in addressing the highly ill-posed problem of large-angle viewpoint changes. After recovering the completed geometry and appearance, we propose an effective alignment strategy to seamlessly integrate the completed 3D structure with the original, ensuring a more coherent and accurate reconstruction. Extensive experiments demonstrate that our method enables zero-shot novel view synthesis with both appearance and geometric consistency across multiple datasets.

We summarize our key contributions as follows:
\begin{itemize}
    \item We propose SceneCompleter, a geometry-aware framework for generative novel view synthesis from sparse input views. It reformulates view generation as dense 3D scene completion, allowing the model to reason about missing geometry and appearance jointly and thereby reducing geometry distortion and appearance drift across generated views.
    \item We develop a geometry-aware generative architecture that unifies RGBD completion, scene-level conditioning, and 3D scene integration. This design enables SceneCompleter to synthesize missing appearance and geometry in a spatially aligned latent space, preserve holistic scene coherence through global context, and convert generated RGBD outputs into an expandable 3D scene representation for iterative completion.
    \item We validate the proposed formulation and architecture through comprehensive experiments on DL3DV-10K, RealEstate10K, Tanks-and-Temples, and CO3D, covering both in-domain and out-of-distribution evaluation. The results demonstrate consistent improvements in visual realism, cross-view coherence, and 3D geometric consistency over regression-based and RGB-only generative baselines.
\end{itemize}

\section{Related Work}

\textbf{Regression-based Novel View Synthesis.}
Conventional approaches in Novel View Synthesis (NVS) rely on dense multi-view images as inputs to learn the 3D representation of the target scene in an optimization manner~\cite{mildenhall2021nerf,huang20242dgs,muller2022ingp}, suffering from inefficiency and limited applicability in practical scenarios.
Sparse-view variants reduce input requirements through feed-forward scene priors, regularization, and geometric priors~\cite{yu2021pixelnerf,niemeyer2022regnerf,yang2023freenerf,wang2023sparsenerf}, but they still reconstruct radiance fields from observed views rather than explicitly completing unseen 3D structure.
Robust variants further relax strict calibration assumptions by jointly refining camera poses and neural fields under imperfect pose estimates~\cite{fu2024cbarf}, but they remain tied to observed-view reconstruction.
Leveraging the fast rendering speed of 3D Gaussians~\cite{kerbl20233dgs}, a series of works~\cite{charatan2024pixelsplat, chen2024mvsplat, fei2024pixelgaussian, wewer2024latentsplat} have shifted toward a new regression-based pipeline, which extracts scene priors from large datasets. These methods directly regress pixel-aligned 3D Gaussian parameters from sparse input views, demonstrating excellent interpolation results and high efficiency when synthesizing novel views close to the reference view. However, the ill-posed nature of this approach causes significant issues when dealing with large viewpoint changes, leading to unrealistic results. In this paper, we explore generative novel view synthesis by using generative models to complete the missing 3D structure in sparse views, thus enabling realistic novel view synthesis even with large viewpoint changes.

\textbf{Generative Novel View Synthesis.}
The impressive capabilities of recent generative models~\cite{ho2020ddpm, rombach2022highstablediffusion} suggest a natural approach to addressing missing viewpoints in novel view synthesis.  
Earlier generative view synthesis methods also explored self-consistent generation for synthesizing novel views~\cite{liu2022deepviewsynthesis}.
Pioneered by Zero-1-to-3~\cite{liu2023zero123}, researchers have begun exploring generative models~\cite{sargent2024zeronvs, long2024wonder3d, liu2023one12345, liu2023syncdreamer, shi2023mvdream, shi2023zero123pp} for object novel view synthesis, framing it as a conditional generation task. MotionCtrl~\cite{wang2024motionctrl} and CameraCtrl~\cite{he2024cameractrl} extend this idea by synthesizing novel views through video generation models, taking a reference image and a camera trajectory as input to generate a sequence of novel view videos. However, since the image itself lacks scale information, the model struggles to learn accurate camera trajectories, making it difficult to generate perspective-correct novel views.
Complementary efforts introduce residual encoders or GAN priors for monocular and degraded-input view synthesis~\cite{guo2026ree3d, guo2025ganprior}, and video diffusion priors further support long-range perpetual view generation~\cite{pan2026dreamjourney}.
Recently, ViewCrafter~\cite{yu2024viewcraftertamingvideodiffusion} and ReconX~\cite{liu2024reconxreconstructscenesparse} have addressed the scale ambiguity by leveraging video generation for novel view synthesis. They project the reference view onto the target view before generation, inherently resolving the scale issue. However, these methods still focus on RGB image completion while weakly constraining missing geometry, which can lead to inconsistent 3D structures. In this paper, we explore 3D generative novel view synthesis by jointly modeling geometry and appearance for dense 3D scene completion, enabling 3D-consistent novel view synthesis.

\textbf{Dense 3D Scene Reconstruction.}
Based on the powerful 3D pointmap representation, Dust3R~\cite{wang2024dust3r} initiates a trend toward dense 3D scene reconstruction without relying on traditional camera models. 
Subsequently, Mast3R~\cite{leroy2024mast3r} further introduces a local feature for each pixel-aligned point to achieve a better performance. 
These works require post-processing and global alignment when reconstructing dense scenes from multiple inputs, as they only establish spatial matching between two views.
To overcome this, Spann3r~\cite{wang2024span3r} and Fast3r~\cite{yang2025fast3r} achieve dense 3D scene reconstruction directly from multiple images by maintaining a spatial memory of past frames or further expanding the original network. 
However, these works only operate in a discriminative manner, densely reconstructing the scene based on as many visual inputs as possible. 
In this paper, we guide this dense reconstruction network to provide unified geometry clues for generative 3D scene completion and then achieve the ultimate novel view synthesis.

\section{Proposed Approach}
\begin{figure*}[t]
    \centering
    \includegraphics[width=1\linewidth]{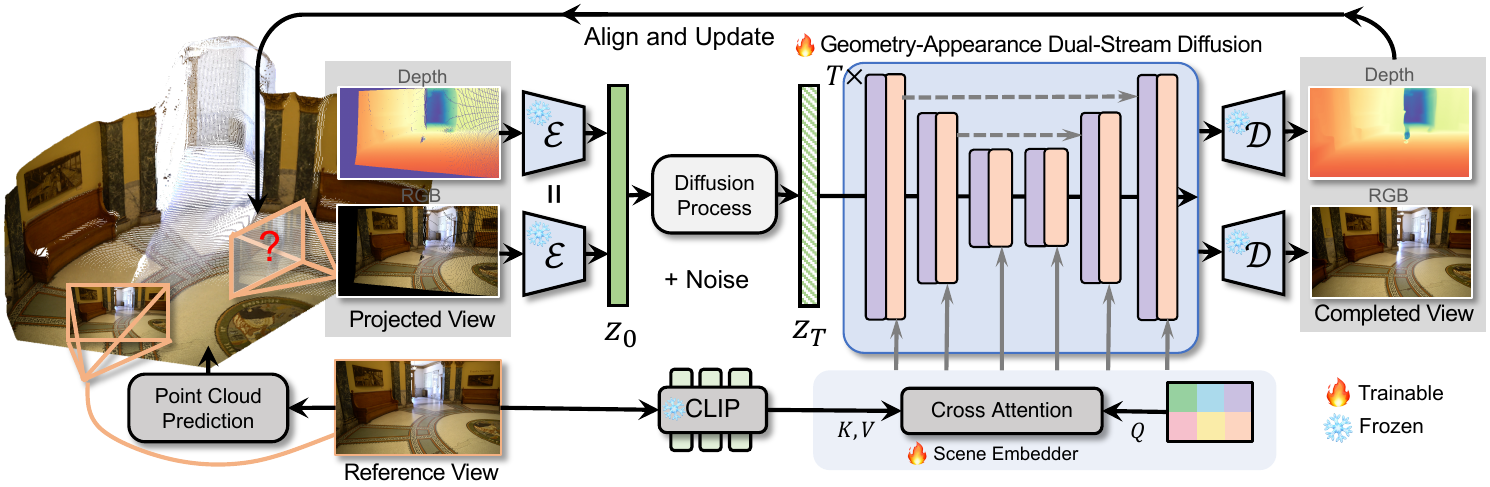}
    \caption{\textbf{Framework of SceneCompleter.} We first extract the geometry-appearance clues from the reference view using an unconstrained stereo reconstruction method. Then, we employ a Geometry-Appearance Dual-Stream Diffusion model to generate novel views in 3D space, conditioned on the extracted geometry-appearance clues. After generating the 3D novel view, we align the synthesized geometry with the original 3D structure to achieve 3D scene completion. Notably, this process can be iterated to progressively generate a larger 3D scene.}
    \label{fig:method}
\end{figure*}

In this section, we introduce \textbf{SceneCompleter}, a novel framework that leverages dense 3D scene completion to achieve geometrically consistent generative novel view synthesis. We first outline the motivation behind SceneCompleter in Section~\ref{sec:motivation}. Then, we describe the \textit{Geometry-Appearance Clue Extraction} process in Section~\ref{sec:clue_extraction}. Next, we introduce the \textit{Scene Embedder} in Section~\ref{sec:scene_embedder} and \textit{Geometry-Appearance Dual-Stream Diffusion} module in Section~\ref{sec:dual_stream}. Finally, we present our \textit{Geometry Alignment and Scene Completion} strategy in Section~\ref{sec:scene_completion}. An overview of our framework is illustrated in Figure~\ref{fig:method}.

\subsection{Motivation of Our SceneCompleter}
\label{sec:motivation}
Recent methods~\cite{yu2024viewcraftertamingvideodiffusion, liu2024reconxreconstructscenesparse} attempt to formulate Novel View Synthesis (NVS) as a 2D inpainting problem by leveraging strong pretrained 2D generative models. While this strategy often produces visually appealing results and alleviates the ill-posedness of sparse-view reconstruction, the synthesized novel views frequently exhibit geometry distortion and appearance drift. The core issue is that such approaches ignore the fundamental nature of NVS: (1) \textbf{cross-view consistency}, i.e., aligning with known pixels and maintaining coherence across generated regions; and (2) \textbf{geometric correctness}, i.e., ensuring that synthesized content respects the 3D structure defined by camera poses. These requirements make NVS inherently a 3D generation problem, which 2D inpainting formulations cannot faithfully satisfy.

To address this limitation, we formulate NVS as a \textbf{3D scene completion} task. Rather than performing 2D inpainting, we aim to infer a complete scene representation by modeling the conditional distribution
\[
p(A, G \mid G_{\text{obs}}, A_{\text{obs}}, F),
\]
where $G_{\text{obs}}$ and $A_{\text{obs}}$ denote the observed geometry clues and appearance clues extracted from input views, and $F$ represents global feature-level information distilled from those observations (e.g., semantic or style cues). The generated appearance $A$ and geometry $G$ provide a single 3D explanation that inherently enforces multi-view consistency and geometric correctness across novel viewpoints.

\subsection{Geometry-Appearance Clue Extraction}
\label{sec:clue_extraction}
To establish the inpainting conditions required by our formulation, we derive the observed geometry and appearance clues $G_{\text{obs}}$ and $A_{\text{obs}}$ from the known views. We employ the unconstrained stereo 3D reconstruction method Dust3R~\cite{wang2024dust3r} to recover dense geometry from the input images. The reconstructed RGB point cloud is then projected into each target camera, producing the appearance clues $A_{\text{obs}}$.
For geometry clues, a straightforward idea is to use Dust3R’s predicted pointmaps. However, in the context of novel view synthesis, the geometry of each target pixel is constrained by the camera parameter, leaving only \textit{one valid degree of freedom} along its corresponding camera ray. Introducing full 3D pointmap values injects redundant degrees of freedom that may break geometric consistency at unseen viewpoints. To avoid this issue, we replace the point map with a projected depth map, which removes unnecessary degrees of freedom. Moreover, depth maps are spatially smoother and structurally more similar to RGB images, making them easier for the model to learn jointly with appearance information. 
The entire process can be formally expressed as:
\[
(G_{\text{obs}}, A_{\text{obs}}) = \text{Proj}(\mathbf{X}_s, I_s, \Pi_t),
\]
where $\mathbf{X}_s$ denotes the 3D points reconstructed from the source view by Dust3R~\cite{wang2024dust3r}, $I_s$ is the corresponding source image, and $\Pi_t$ represents the target camera projection. $\text{Proj}$ extracts the depth as $G_{\text{obs}}$ and the RGB as $A_{\text{obs}}$.

\subsection{Scene Embedder}
\label{sec:scene_embedder}
To further mitigate geometry distortion and appearance drift and enforce cross-view consistency, we design a \textbf{Scene Embedder} that provides global context from reference images. Inspired by Q-Former~\cite{li2023blipQformer}, it leverages a learnable scene embedding to interact with features extracted from a reference view, encoding holistic scene information. Specifically, we extract reference features $f_{\text{ref}}$ using a pretrained CLIP image encoder~\cite{radford2021learningclip}, and let the learnable embedding $f_{\text{emb}}$ query these features through cross-attention:
\[
f_{\text{scene}} = \text{CrossAttn}\Big(W_q(f_{\text{emb}}), W_k(f_{\text{ref}}), W_v(f_{\text{ref}})\Big),
\]
where $W_q$, $W_k$, and $W_v$ are the query, key, and value projections. The resulting scene embedding $f_{\text{scene}}$ is incorporated into the U-Net via cross-attention, effectively injecting global scene context to guide generation in missing regions.
With this design, the Scene Embedder enforces consistent global context across all views, providing guidance to the model in unobserved regions and alleviating the inherent ill-posedness of novel view synthesis, while remaining lightweight and efficiently integrated via cross-attention without requiring full image-level processing.

\subsection{Geometry-Appearance Dual-stream Diffusion}
\label{sec:dual_stream}
Our goal is to achieve dense 3D scene completion, where the key challenge lies in effectively incorporating geometry $G$ into the completion process while preserving the generative model’s capability for image completion. To address this, we design a depth encoder-decoder that shares weights with the image encoder, mapping the depth map into a latent space. The latent geometry is fused with the image latent and processed by a single Geometry-Appearance dual-stream U-Net to simultaneously complete both appearance and geometry in the latent space. Our model builds upon Stable Diffusion 2~\cite{rombach2022highstablediffusion}, pretrained on the large-scale LAION-5B~\cite{schuhmann2022laion5b} dataset, leveraging its strong priors for natural image generation.

\textbf{Depth encoder and decoder.}
A straightforward way to encode both depth and image into the latent space is to modify the channel dimensions of the diffusion model's VAE. However, this requires reconstructing the entire latent space, which would compromise the pretrained priors learned by the diffusion model for natural images. Instead, we employ two separate VAEs to encode depth and image independently. For the depth VAE, we replicate the depth map three times to satisfy the VAE's three-channel input requirement. To account for numerical scale differences between images and depth maps, we apply an affine-invariant depth normalization:
\begin{equation}
    d_n = \left( \frac{d - d_{2}}{d_{98} - d_{2}} - 0.5 \right) \times 2,
\end{equation}
where $d$ and $d_n$ are the depth maps before and after normalization, and $d_{2}$ and $d_{98}$ are the 2nd and 98th percentile values of the depth map. Similar to Marigold~\cite{ke2024repurposingmarigold}, this ensures the original SD-VAE can reconstruct depth losslessly. Accordingly, we use two weight-sharing VAEs to encode and decode depth and image independently without any fine-tuning:
\begin{gather}
    z_d = \mathcal{E}(d_n), \quad \hat{d} = \mathcal{D}(z_d), \\
    z_i = \mathcal{E}(I), \quad \hat{I} = \mathcal{D}(z_i),
\end{gather}
where $z_d$ and $z_i$ are the latent codes for depth and image, $\mathcal{E}$ and $\mathcal{D}$ denote the encoder and decoder, and $\hat{d}$ and $\hat{I}$ are the reconstructed depth and image.

\textbf{Geometry-appearance denoising U-Net.}  
To simultaneously complete geometry and appearance $(A, G)$, we perform generation in the 3D latent space of image and depth. A key challenge is how to construct the latent representation. We choose to \textbf{concatenate the geometry and appearance latents along the channel dimension}, rather than treating them as separate latents or concatenating spatially into a large image. This design inherently aligns the spatial positions of the two latents, so that each feature corresponds to a 3D feature.
We apply the same concatenation strategy to the \textbf{validity mask}, which indicates whether a feature at a given position originates from input views or should be completed. This mask is obtained from the projection-based feature interpolation. Consequently, the initial latent is constructed by concatenating the validity mask, the projected and ground-truth depth and image latents:
\[
z_0 = \text{concat}(m, z_d^{\text{proj}}, z_i^{\text{proj}}, z_d^{\text{gt}}, z_i^{\text{gt}}),
\]
where $m$ is the validity mask, $z_d^{\text{proj}}$ and $z_i^{\text{proj}}$ are the depth and image latents from observed (projected) regions, and $z_d^{\text{gt}}$ and $z_i^{\text{gt}}$ are the corresponding ground-truth latents.

During the reverse diffusion process, the U-Net predicts the noise $\hat{\epsilon}$ conditioned on the noised latent $z_c$, the initial latent $z_0$, the scene embedding $f_\text{scene}$, and the timestep $t$. The training loss is formulated as:
\begin{equation}
\mathcal{L} = \mathbb{E}_{d_0, \epsilon \sim \mathcal{N}(0, 1), t \sim U(1, T)} \Big[ \big\| \epsilon - \hat{\epsilon}_\theta(z_c, z_0, t, f_\text{scene}) \big\|^2 \Big].
\end{equation}
This design ensures that the model jointly completes appearance and geometry in a spatially aligned 3D latent space while leveraging global scene context.

\subsection{Geometry Alignment and 3D Scene Completion}
\label{sec:scene_completion}
Since we apply affine-invariant normalization during 3D scene completion, it is necessary to align the completed scene with the original geometry. This alignment can be achieved by matching the predicted depth \( \hat{d} \) with the incomplete depth clues \( d_p \). Specifically, we use the valid depth mask \( M_{d_p} \) to locate the corresponding predicted depths \( \hat{d}_p \), and then apply a least-squares fitting to align \( \hat{d}_p \) to \( d_p \). 
The optimization for computing the scale and offset is:
\begin{equation}
s, o = \arg\min_{s, o} \| d_p - (s \hat{d}_p + o) \|^2.
\end{equation}
Here, \(s\) and \(o\) denote the scale and offset that map the normalized predicted depth to the coordinate scale of the projected depth clues. We then apply them to the full predicted depth map to restore the missing areas in the scene. The final aligned depth $\hat{d}_{\text{aligned}}$ is yielded as follows:
\begin{equation}
    \hat{d}_{\text{aligned}} = s \hat{d} + o.
\end{equation}
After obtaining \( \hat{d}_{\text{aligned}} \), we can restore it to a 3D pointmap $\mathbf{\hat{X}}_i$ using the camera parameters \( K_i \), \( R_i \), and \( T_i \) as follows:
\begin{equation}
    \mathbf{\hat{X}}_i = R_i^{-1} K_i^{-1} \tilde{\mathbf{p}}_i \hat{d}_{\text{aligned}} - R_i^{-1} T_i,
\end{equation}
where \( \tilde{\mathbf{p}}_i \) denotes the pixel homogeneous coordinates.

\section{Experiments}
\subsection{Implementation Details}
Our model is built upon Stable Diffusion v2~\cite{rombach2022highstablediffusion}. Following ViewCrafter~\cite{yu2024viewcraftertamingvideodiffusion}, we train our model using the DL3DV-10K~\cite{ling2024dl3dv} and RealEstate10K~\cite{zhou2018stereore10k} datasets. Since neither dataset provides calibrated depth or other geometric information, we employ Dust3R~\cite{wang2024dust3r} to generate the required training data. Specifically, we randomly sample five frames with temporal strides of 1, 2, 4, and 8 to simulate diverse viewpoint changes, and estimate their depth maps, camera poses, and intrinsics with Dust3R. During training, each target frame is paired with a randomly projected source frame as the conditional input. We train the model for 50k iterations with an effective batch size of 32 and a learning rate of 3e-5 on 8 NVIDIA A6000 GPUs.
\subsection{Zero-shot Novel View Synthesis Comparison}
\textbf{Datasets and Metrics}
We evaluate our model's generative novel view synthesis on the zero-shot DL3DV-10K~\cite{ling2024dl3dv}, RealEstate10K~\cite{zhou2018stereore10k} test sets, and out-of-distribution datasets Tanks-and-Temples~\cite{knapitsch2017tankstnt} and CO3D~\cite{reizenstein2021commonco3d}. We choose the regression-based method Dust3R, the 2D generative-based methods MotionCtrl, and ViewCrafter as our baselines. For 2D metrics, we use PSNR, SSIM, and LPIPS for evaluation. For 3D metrics, we calculate the camera rotation distance \( R_{\text{dist}} \) and translation distance \( T_{\text{dist}} \) as follows:
\begin{gather}
    \label{eq:r_dist}
    R_{\text{dist}} = \sum_{i=1}^n \arccos({\frac{\text{tr}(\mathbf{R}_{\text{gen}}^{i} \mathbf{R}_{\text{gt}}^{i\mathrm{T}}) - 1}{2}}),
\end{gather}
\begin{equation}
     \label{eq:t_dist}
     T_{\text{dist}} = \sum_{i=1}^n \|\mathbf{T}_{\text{gt}}^{i} - \mathbf{T}_{\text{gen}}^{i} \|_2,
 \end{equation}
where \( \mathbf{R}_{\text{gen}}^i \) and \( \mathbf{R}_{\text{gt}}^i \) are the predicted and ground truth rotation matrices, respectively, and \( \mathbf{T}_{\text{gen}}^i \) and \( \mathbf{T}_{\text{gt}}^i \) are the predicted and ground truth translation vectors, respectively.

\begin{table*}[t] \small
\caption{\textbf{Quantitative comparison of zero-shot novel view synthesis on Tanks-and-Temples\cite{knapitsch2017tankstnt}, RealEstate10K\cite{zhou2018stereore10k}, DL3DV-10K~\cite{ling2024dl3dv} and CO3D\cite{reizenstein2021commonco3d} datasets.}
Our SceneCompleter outperforms baselines across most image quality and pose accuracy metrics.
}
\label{tab:comp_zero}
\centering
\begin{tabular}{lcccccccccc}
\cmidrule[\heavyrulewidth]{1-11}
 \textbf{Dataset}
 & \multicolumn{5}{c}{\textbf{Easy set}} & \multicolumn{5}{c}{\textbf{Hard set}} \\
 \cmidrule(lr){2-6} \cmidrule(lr){7-11}
 Method & LPIPS $\downarrow$ & PSNR $\uparrow$ & SSIM $\uparrow$  & $R_{\text{dist}}$ $\downarrow$ & $T_{\text{dist}}$ $\downarrow$ & LPIPS $\downarrow$ & PSNR $\uparrow$ & SSIM $\uparrow$ & $R_{\text{dist}}$ $\downarrow$ & $T_{\text{dist}}$ $\downarrow$\\ \cmidrule{1-11}
\textbf{Tanks-and-Temples}\\
Dust3R~\cite{wang2024dust3r}&0.478&16.26&\cellcolor{tabthird}0.506&\cellcolor{tabsecond}0.173&1.021 &0.527&14.74&0.368&\cellcolor{tabsecond}0.498&\cellcolor{tabsecond}1.125\\
MotionCtrl~\cite{wang2024motionctrl}&\cellcolor{tabthird}0.415&\cellcolor{tabthird}16.55&0.498&0.222&\cellcolor{tabthird}0.992&\cellcolor{tabthird}0.464&\cellcolor{tabthird}15.52&\cellcolor{tabthird}0.437&0.578&1.384\\
ViewCrafter~\cite{yu2024viewcraftertamingvideodiffusion}&\cellcolor{tabsecond}0.217&\cellcolor{tabsecond}20.67&\cellcolor{tabsecond}0.668&\cellcolor{tabthird}0.213&\cellcolor{tabsecond}0.853&\cellcolor{tabsecond}0.273&\cellcolor{tabsecond}18.50&\cellcolor{tabsecond}0.554&\cellcolor{tabthird}0.514&\cellcolor{tabthird}1.200\\
Ours&\cellcolor{tabfirst}0.207&\cellcolor{tabfirst}21.43&\cellcolor{tabfirst}0.700&\cellcolor{tabfirst}0.163&\cellcolor{tabfirst}0.828&\cellcolor{tabfirst}0.247&\cellcolor{tabfirst}19.80&\cellcolor{tabfirst}0.555&\cellcolor{tabfirst}0.496&\cellcolor{tabfirst}1.037\\
\midrule
\textbf{RealEstate10K}\\
Dust3R~\cite{wang2024dust3r}&0.689&12.55&0.496&\cellcolor{tabthird}0.046&\cellcolor{tabthird}0.174&0.661&12.31&0.490&\cellcolor{tabthird}0.047&\cellcolor{tabthird}0.169\\
MotionCtrl~\cite{wang2024motionctrl}&\cellcolor{tabfirst}0.102&\cellcolor{tabsecond}22.88&\cellcolor{tabsecond}0.810&0.116&1.937&\cellcolor{tabfirst}0.117&\cellcolor{tabsecond}22.56&\cellcolor{tabsecond}0.808&\cellcolor{tabsecond}0.031&1.051\\
ViewCrafter~\cite{yu2024viewcraftertamingvideodiffusion}&\cellcolor{tabthird}0.141&\cellcolor{tabthird}22.43&\cellcolor{tabthird}0.807&\cellcolor{tabfirst}0.021&\cellcolor{tabsecond}0.134&\cellcolor{tabthird}0.161&\cellcolor{tabthird}22.01&\cellcolor{tabthird}0.802&\cellcolor{tabfirst}0.030&\cellcolor{tabsecond}0.149\\
Ours&\cellcolor{tabsecond}0.121&\cellcolor{tabfirst}26.03&\cellcolor{tabfirst}0.867&\cellcolor{tabsecond}0.035&\cellcolor{tabfirst}0.121&\cellcolor{tabsecond}0.118&\cellcolor{tabfirst}25.94&\cellcolor{tabfirst}0.868&\cellcolor{tabsecond}0.031&\cellcolor{tabfirst}0.143\\
\midrule
\textbf{DL3DV-10K}\\
Dust3R~\cite{wang2024dust3r}&0.660&13.63&0.429&\cellcolor{tabthird}0.828&1.181&0.741&10.05&\cellcolor{tabsecond}0.488&\cellcolor{tabsecond}0.819&\cellcolor{tabsecond}0.785\\
MotionCtrl~\cite{wang2024motionctrl}&\cellcolor{tabthird}0.540&\cellcolor{tabthird}16.74&\cellcolor{tabthird}0.657&\cellcolor{tabsecond}0.412&\cellcolor{tabthird}1.107&\cellcolor{tabthird}0.585&\cellcolor{tabthird}14.90&0.462&0.822&\cellcolor{tabthird}0.951\\
ViewCrafter~\cite{yu2024viewcraftertamingvideodiffusion}&\cellcolor{tabsecond}0.346&\cellcolor{tabsecond}22.91&\cellcolor{tabsecond}0.697&2.215&\cellcolor{tabsecond}1.098&\cellcolor{tabsecond}0.426&\cellcolor{tabsecond}18.49&\cellcolor{tabthird}0.472&\cellcolor{tabthird}0.821&\cellcolor{tabthird}0.951\\
Ours&\cellcolor{tabfirst}0.192&\cellcolor{tabfirst}24.38&\cellcolor{tabfirst}0.789&\cellcolor{tabfirst}0.369&\cellcolor{tabfirst}0.456&\cellcolor{tabfirst}0.271&\cellcolor{tabfirst}21.25&\cellcolor{tabfirst}0.660&\cellcolor{tabfirst}0.368&\cellcolor{tabfirst}0.640\\
\midrule
\textbf{CO3D}\\
Dust3R~\cite{wang2024dust3r}&0.555&\cellcolor{tabthird}13.40&\cellcolor{tabsecond}0.284&\cellcolor{tabfirst}0.163&1.363&0.595&8.67&\cellcolor{tabsecond}0.257&\cellcolor{tabsecond}2.334&\cellcolor{tabthird}1.779\\
MotionCtrl~\cite{wang2024motionctrl}&\cellcolor{tabthird}0.531&11.03&0.147&\cellcolor{tabthird}0.171&\cellcolor{tabthird}1.214&\cellcolor{tabsecond}0.394&\cellcolor{tabthird}11.64&\cellcolor{tabthird}0.178&\cellcolor{tabthird}2.607&\cellcolor{tabsecond}0.968\\
ViewCrafter~\cite{yu2024viewcraftertamingvideodiffusion}&\cellcolor{tabsecond}0.399&\cellcolor{tabsecond}15.14&\cellcolor{tabthird}0.263&0.178&\cellcolor{tabsecond}1.197&\cellcolor{tabthird}0.548&\cellcolor{tabsecond}14.54&0.121&2.610&\cellcolor{tabsecond}0.968\\
Ours&\cellcolor{tabfirst}0.378&\cellcolor{tabfirst}17.45&\cellcolor{tabfirst}0.326&\cellcolor{tabsecond}0.168&\cellcolor{tabfirst}0.607&\cellcolor{tabfirst}0.374&\cellcolor{tabfirst}15.07&\cellcolor{tabfirst}0.306&\cellcolor{tabfirst}2.330&\cellcolor{tabfirst}0.607\\
\bottomrule
\end{tabular}
\end{table*}

\begin{figure*}[t]
    \centering
    \includegraphics[width=1\linewidth]{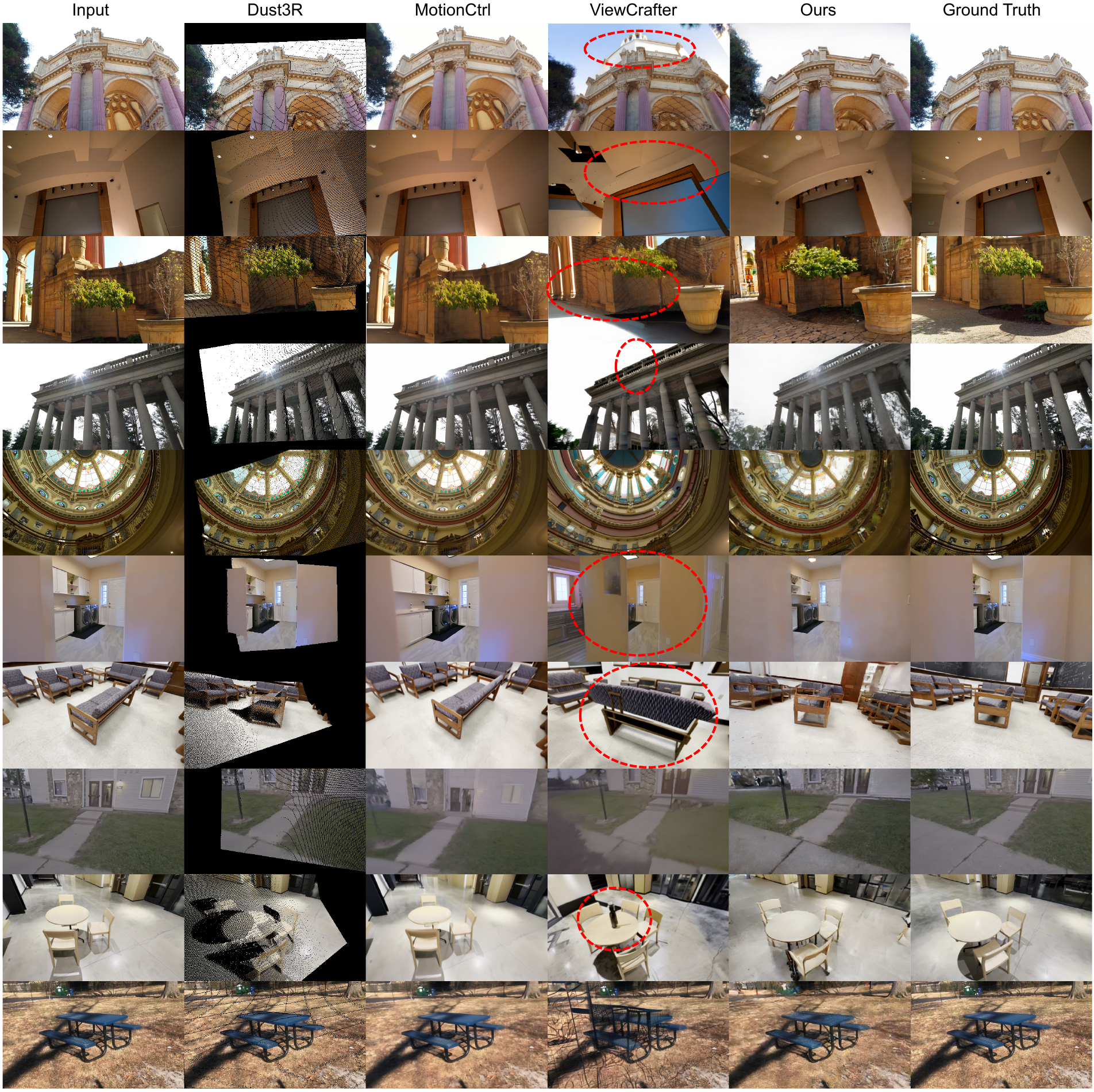}
    \caption{\textbf{Qualitative comparison of zero-shot novel view synthesis on Tanks-and-Temples~\cite{knapitsch2017tankstnt}, RealEstate10K~\cite{zhou2018stereore10k}, DL3DV-10K~\cite{ling2024dl3dv}, CO3D~\cite{reizenstein2021commonco3d} datasets.} Our SceneCompleter achieves more realistic and 3D-consistent novel view synthesis. }
    \label{fig:comparison_main}
\end{figure*}

\textbf{Quantitative Comparison.}
Table~\ref{tab:comp_zero} presents our quantitative comparison experiments, where we consider both 2D appearance and 3D structure metrics. Additionally, we divide the dataset into an easy set and a hard set for more granular comparisons. In terms of 2D metrics, our model outperforms all others in PSNR and SSIM, primarily due to our joint modeling of both geometry and appearance, which results in more consistent and realistic structures in novel view synthesis. Additionally, we achieve the best LPIPS scores in most cases (except for one), demonstrating that our model not only generates high-quality and geometrically consistent structures but also produces visually realistic appearances in novel view synthesis. In terms of 3D metrics, our method consistently outperforms 2D-based generative methods, thanks to the incorporation of geometry information into the generation process, which ensures more consistent structures. However, our 3D metrics sometimes fall slightly behind the regression-based method, Dust3R. This could be due to some details being filled in the missing areas during generation, which may influence the calculation of 3D metrics.

\textbf{Qualitative comparison.}
Figure~\ref{fig:comparison_main} shows the qualitative comparison results of our method. Dust3R~\cite{wang2024dust3r}, being a regression-based method, lacks generative capabilities, resulting in missing areas in the novel view generation. MotionCtrl~\cite{wang2024motionctrl}, on the other hand, synthesizes novel views based on a single image and camera trajectory. However, the image itself lacks scale information, leading to a mismatch between the camera trajectory's scale and the image scale, which makes it difficult to control the new viewpoint. As shown in Figure~\ref{fig:comparison_main}, MotionCtrl often exhibits minimal camera viewpoint changes, leading to inaccurate novel view synthesis. ViewCrafter~\cite{yu2024viewcraftertamingvideodiffusion} projects the reference view to the novel view and relies on a video generation model to complete the overall image, naturally avoiding scale issues by directly using camera projection to obtain the incomplete conditional image. However, by relying solely on RGB input without 3D information, ViewCrafter sometimes struggles to understand scene relationships, resulting in incorrect scene results or the addition/removal of scene content. For example, in the second row, the generated 3D structure appears inconsistent, while in the second-to-last row, an extra wine bottle is erroneously added to the table, conflicting with other viewpoints. Our SceneCompleter, which simultaneously models both geometry and appearance, benefits from structural guidance, ensuring superior 3D consistency. Additionally, our Scene Embedder encodes global scene information, enabling our model to effectively complete large missing regions while preserving consistency with the original structure, even under significant camera viewpoint changes.

\textbf{Generalization analysis.}
Beyond in-domain evaluation on DL3DV-10K and RealEstate10K, we explicitly test SceneCompleter on the out-of-distribution Tanks-and-Temples and CO3D datasets. As shown in Table~\ref{tab:comp_zero} and Figure~\ref{fig:comparison_main}, our method maintains strong visual quality and 3D consistency across both scene-level and object-centric data. The improvements are especially clear on hard splits and out-of-distribution scenes, where RGB-only generative baselines often hallucinate inconsistent content and regression-based reconstruction leaves large missing regions. These results indicate that modeling geometry and appearance jointly provides more robust scene priors for large-viewpoint generative novel view synthesis.

\subsection{3D Scene Completion}
\begin{figure*}[t]
    \centering
    \includegraphics[width=1\linewidth]{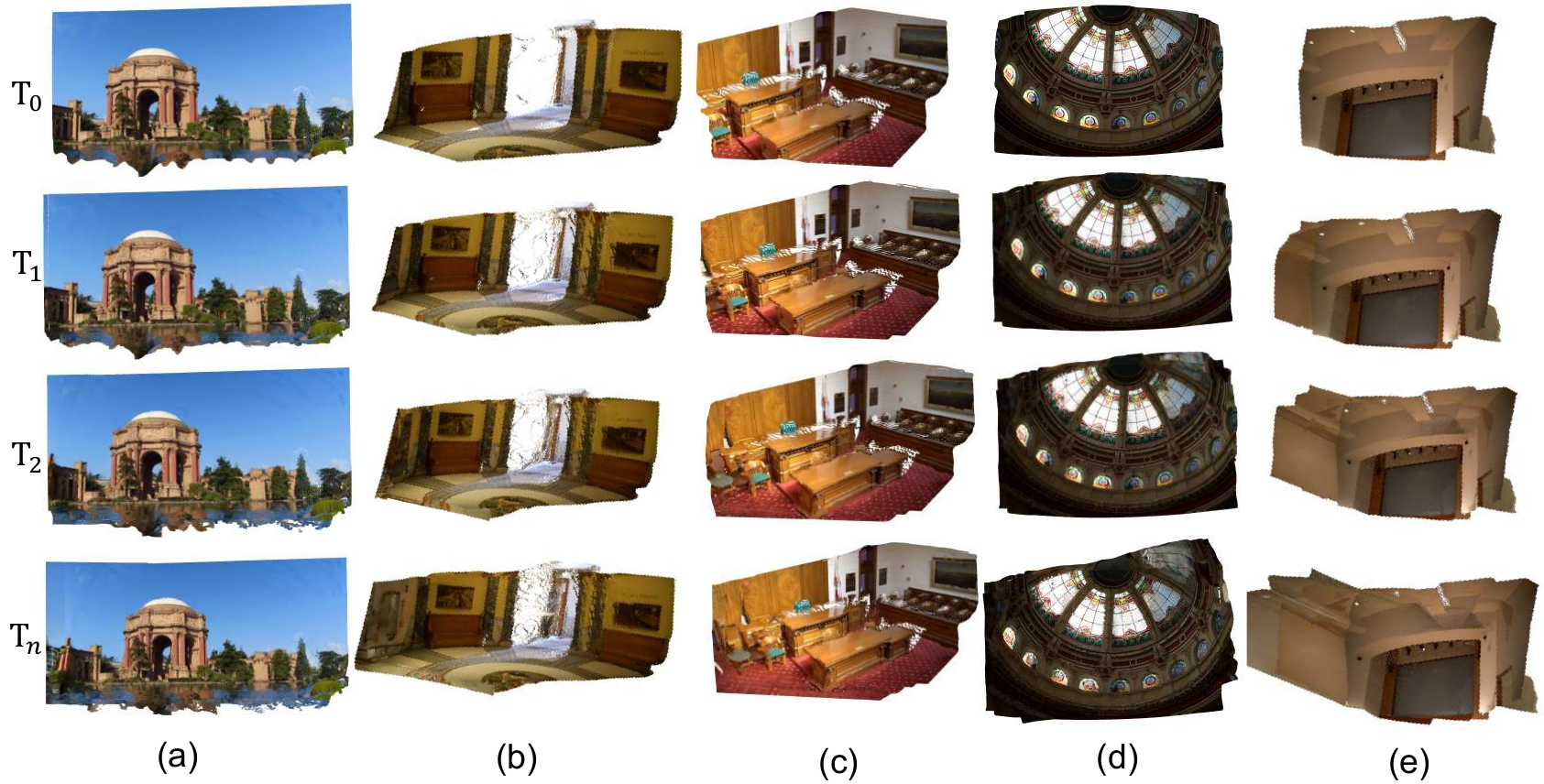}
    \caption{\textbf{3D scene completion visualization.} Our SceneCompleter can iteratively refine 3D scenes. }
    \label{fig:3d_vis}
\end{figure*}

We present our 3D scene completion results in Figure~\ref{fig:3d_vis}. From left to right, we iteratively update and refine the scene information, gradually improving the completion. As shown, leveraging our joint modeling of geometry and appearance, along with our simple yet effective alignment strategy, our model achieves iterative and coherent 3D scene completion while preserving the original 3D structure, enabling single-image 3D scene generation. Notably, as illustrated in Figure~\ref{fig:3d_vis}(d), our model adapts not only to camera translation but also to camera rotation, demonstrating its strong robustness.

\subsection{Ablation Study}
\begin{figure}[t]
    \centering
    \includegraphics[width=1\linewidth]{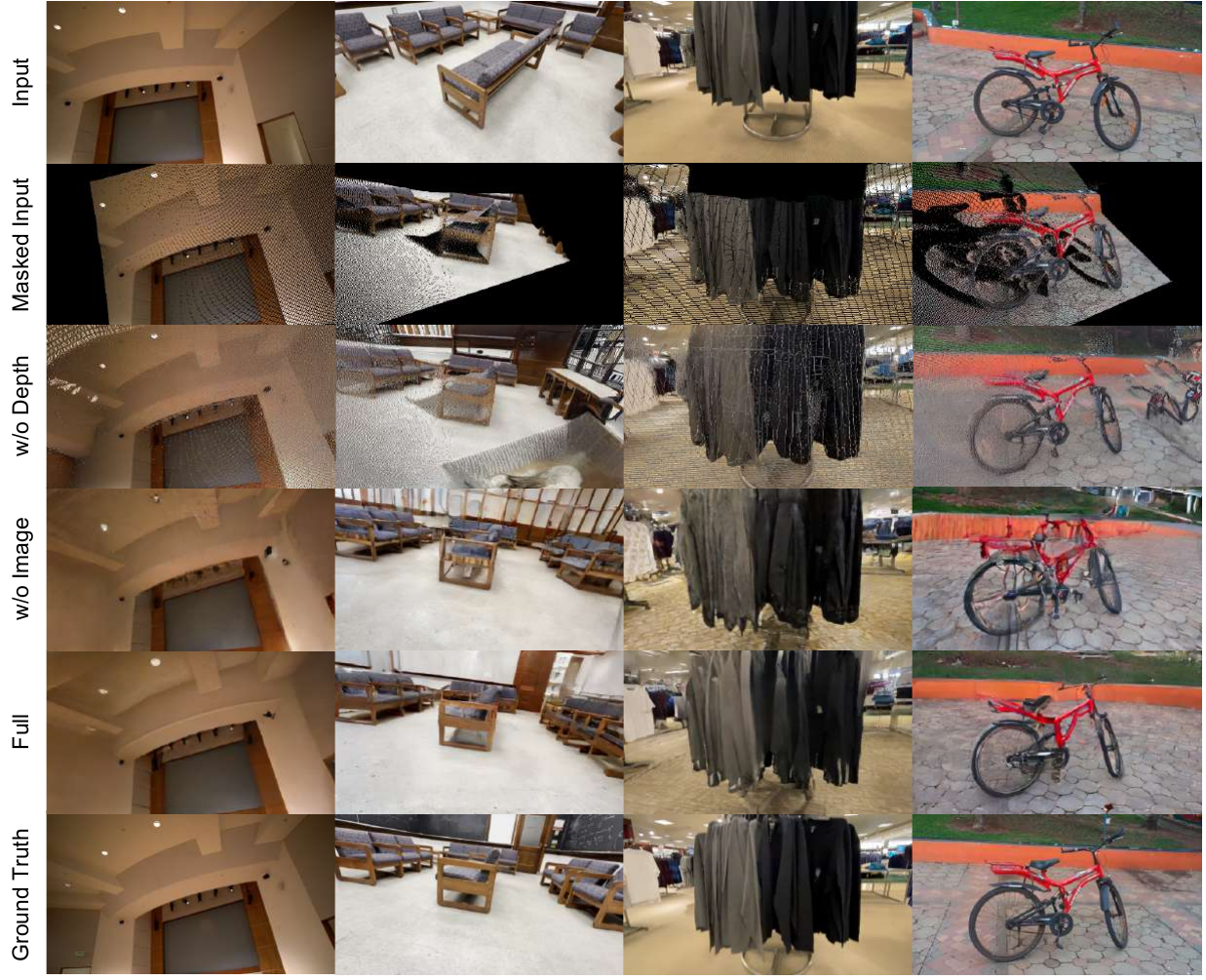}
    \caption{\textbf{Ablation on geometry clue and Scene Embedder}. The results demonstrate that the geometry clues play a decisive role in novel view synthesis, while the scene information encoded by the Scene Embedder is crucial for detail recovery.}
    \label{fig:ablation_vis}
\end{figure}

\begin{table}[t]
\caption{\textbf{Quantitative ablation on geometry clue and Scene Embedder.} The results demonstrate that both the geometry clue and Scene Embedder play crucial roles in the final outcome.
}
\label{tab:comp_ablation}
\centering
\resizebox{1.0\linewidth}{!}{%
\begin{tabular}{lccccc}
\cmidrule[\heavyrulewidth]{1-6}
 \textbf{Dataset} & \multicolumn{5}{c}{\textbf{Test set}} \\
 \cmidrule(lr){2-6} 
 Method & LPIPS $\downarrow$ & PSNR $\uparrow$ & SSIM $\uparrow$  & $R_{\text{dist}}$ $\downarrow$ & $T_{\text{dist}}$ $\downarrow$\\ \cmidrule{1-6}
\textbf{TNT}\\
w/o depth & 0.727 & 18.82 & 0.431 & 2.805 & 0.920 \\
w/o SE & 0.305 & 21.66 & 0.798 & 2.802 & 0.896 \\
Full & \cellcolor{tabfirst}0.275 & \cellcolor{tabfirst}23.19 & \cellcolor{tabfirst}0.826 & \cellcolor{tabfirst}2.790 & \cellcolor{tabfirst}0.752 \\
\midrule
\textbf{Re10K}\\
w/o depth & 0.726 & 10.92 & 0.224 & 0.592 & 1.246 \\
w/o SE & 0.404 & 15.44 & 0.484 & 0.373 & 1.000 \\
Full & \cellcolor{tabfirst}0.391 & \cellcolor{tabfirst}15.56 & \cellcolor{tabfirst}0.510 & \cellcolor{tabfirst}0.369 & \cellcolor{tabfirst}0.752 \\
\midrule
\textbf{DL3DV}\\
w/o depth & 0.810 & 14.83 & 0.128 & 1.993 & 0.960 \\
w/o SE & 0.459 & 18.28 & 0.456 & 2.086 & 0.948 \\
Full & \cellcolor{tabfirst}0.387 & \cellcolor{tabfirst}19.97 & \cellcolor{tabfirst}0.506 & \cellcolor{tabfirst}1.785 & \cellcolor{tabfirst}0.824 \\
\midrule
\textbf{CO3D}\\
w/o depth & 0.586 & 18.20 & 0.277 & 0.387 & 0.794 \\
w/o SE & 0.341 & 18.18 & 0.373 & 0.111 & \cellcolor{tabfirst}0.287 \\
Full & \cellcolor{tabfirst}0.328 & \cellcolor{tabfirst}19.11 & \cellcolor{tabfirst}0.420 & \cellcolor{tabfirst}0.101 & \cellcolor{tabfirst}0.287 \\
\bottomrule

\end{tabular}
}
\end{table}

In this section, we conduct an ablation study to validate the effectiveness of our design. Specifically, we focus on two crucial components: the joint modeling of geometry and appearance, which ensures structural consistency during 3D scene completion, and the global scene embedder, which provides holistic scene understanding for improved completion quality.

\textbf{Quantitative Results.}
Table~\ref{tab:comp_ablation} shows the results of our quantitative ablation study. As observed, after removing the depth guidance, both 2D and 3D metrics decrease, with significant drops in LPIPS and SSIM. This highlights the importance of modeling geometry for achieving reasonable and structured image completion. Additionally, when the Scene Embedder is ablated, both 2D and 3D metrics show some decline, which aligns with our qualitative analysis, where we observed that the Scene Embedder helps the model handle detailed information more effectively.

\textbf{Qualitative Results.}
 Figure~\ref{fig:ablation_vis} presents our qualitative ablation study. From the figure, we observe the following key points: 1) Simultaneously modeling geometry is crucial for accurate geometric structure prediction. For example, in the second row, the image after completion still contains significant noise, likely due to the absence of depth information for flat surfaces. Additionally, in the second column, the lack of geometric structure information results in unreasonable floating shadows in the missing areas, which would be even more evident in the depth map as incorrect. 2) The global scene information encoded by the scene embedder plays a critical role in handling fine details. For instance, in the first column, the beams appear structurally deformed, with their sharp edges lost, and in the fourth row, the bike’s projection information appears messy. When scene information is available, the model can accurately infer the bike's structure, but in its absence, it generates incorrect structures.
 
\begin{figure}[h]
  \centering
  \includegraphics[width=1.0\linewidth]{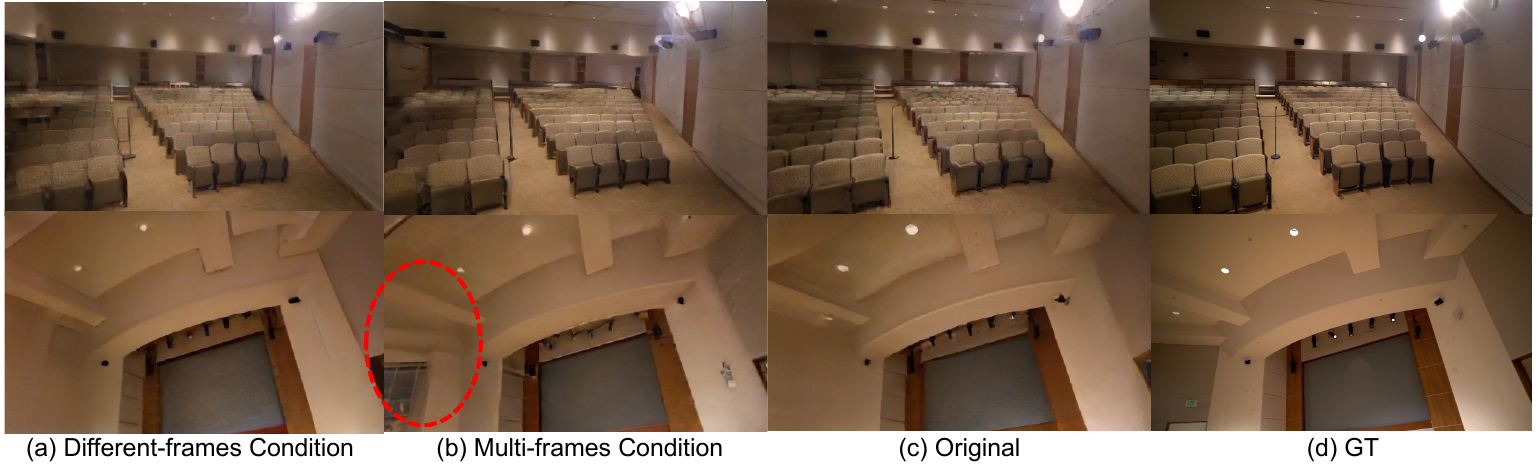}
   \caption{\textbf{Different choices of reference view for the Scene Embedder.} Each single view provides similar global context for the scene, while using multiple reference views can further help infer 3D structure by aggregating complementary information.}
   \label{fig:sceneembedder}
\end{figure}
\subsection{Discussion on Scene Embedder}
Figure~\ref{fig:sceneembedder} analyzes how the choice of reference views affects the Scene Embedder. Since the Scene Embedder is designed to capture global scene-level information rather than local pixel correspondences, different single reference views provide comparable guidance for preserving the overall style, semantics, and layout of the scene. As a result, SceneCompleter remains robust to the specific reference view selected for conditioning and produces consistent completion results across different choices.

We also observe that using multiple reference views can further improve the fidelity of challenging regions. By aggregating complementary observations from different viewpoints, the Scene Embedder obtains a more complete global prior and can better infer structures that are ambiguous from a single image. This behavior confirms that the Scene Embedder does not simply copy local textures from the reference image; instead, it provides scene-level context that helps the diffusion model generate plausible and geometrically coherent missing regions. In our main experiments, we use a randomly selected single reference view for efficiency unless otherwise specified.

\section{Conclusion}

In this paper, we presented SceneCompleter, a geometry-aware framework for generative novel view synthesis from sparse input views. By reformulating view generation as dense 3D scene completion, SceneCompleter jointly reasons about missing geometry and appearance rather than relying on isolated 2D image hallucination. Our geometry-appearance dual-stream diffusion model completes RGB and depth in a spatially aligned latent space, while the Scene Embedder injects holistic scene context to improve global coherence and detail recovery. We further introduced a geometry alignment and integration strategy that converts completed RGBD predictions into an expandable 3D scene representation, enabling iterative scene completion under camera translation and rotation. Experiments across DL3DV-10K, RealEstate10K, Tanks-and-Temples, and CO3D show that explicit geometry-appearance modeling improves visual realism, cross-view coherence, and 3D consistency over regression-based and RGB-only generative baselines.

\bibliographystyle{IEEEtran}
\bibliography{main}

\end{document}